\documentclass{ecai}
\usepackage{times}
\usepackage{graphicx}
\usepackage{latexsym}

\usepackage{amsmath}
\usepackage{amssymb}
\usepackage{siunitx}
\newcommand{\XX}{\mathbf{X}}
\newcommand{\UU}{\mathbf{U}}
\newcommand{\YY}{\mathbf{Y}}
\newcommand{\FF}{\mathbf{F}}

\ecaisubmission   

\begin{document}

\title{Polynomial Neural Networks and Taylor maps for \\
Dynamical Systems Simulation and Learning}

\author{Andrei Ivanov\institute{DESY,
Germany, email: andrei.ivanov@desy.de}
\and Anna Golovkina\institute{Saint Petersburg State University,
Russia, email: a.golovkina@spbu.ru}
\and Uwe Iben\institute{OOO Robert Bosch,
Russia, email: uwe.iben@de.bosch.com}
 }

\maketitle
\bibliographystyle{ecai}

\begin{abstract}
The connection of Taylor maps and polynomial neural networks (PNN) to solve ordinary differential equations (ODEs) numerically is considered.
Having the system of ODEs, it is possible to calculate weights of PNN that simulates the dynamics of these equations. It is shown that proposed PNN architecture can provide better accuracy with less computational time in comparison with traditional numerical solvers. Moreover, neural network derived from the ODEs can be used for simulation of system dynamics with different initial conditions, but without training procedure. On the other hand, if the equations are unknown, the weights of the PNN can be fitted in a data-driven way. In the paper we describe the connection of PNN with differential equations in a theoretical way along with the examples for both dynamics simulation and learning with data.

\end{abstract}

\section{Introduction and Related Works}
\label{sec:1001}
Traditional methods for solving systems of differential equations imply a numerical step-by-step integration of the system. For some problems such as stiff ODEs, this integration procedure leads to time-consuming algorithms because of the limitations on the time step that is used to achieve the necessary accuracy of the solution. From this perspective, neural network as a universal function approximator can be applied for the construction of the solution in a more efficient way. This property induces wide research and consequently large number of papers devoted to the neural networks design for ODEs and PDEs solutions. We go into some of the publications in more detail.  Further, we give a short review and comparison of existing approaches as well as highlight the advantages of proposed in this paper methodology.

In \cite{ref1}, the method to solve initial and boundary value problems using feed-forward neural networks is proposed. The solution of the differential equation is written as a sum of two parts. The first part satisfies the initial/boundary conditions. The second part corresponds to a neural network output. The same technique is applied for solving Stokes problem in \cite{ref2,ref3,ref4}.

In \cite{ref5}, the neural network is trained to satisfy the differential operator, initial condition, and boundary conditions for the partial differential equation (PDE). The authors in \cite{ref6} translate a PDE to a stochastic control problem and use deep reinforcement learning for an approximation of derivative of the solution with respect to the space coordinate.

Other approaches rely on the implementation of a traditional step-by-step integrating method in a neural network framework, cf. \cite{ref7,ref8}. In  \cite{ref8}, the author proposes such an architecture. After fitting, the neural network (NN) produces an optimal finite difference scheme for a specific system. The back-propagation technique through an ordinary differential equation (ODE) solver is proposed in \cite{ref9}. The authors construct a certain type of neural network that is analogous to a discretized differential equation. This group of methods requires a traditional numerical method to simulate dynamics.

Polynomial neural networks are also widely presented in the literature, cf. \cite{ref10,ref11,ref12}. In \cite{ref10}, the polynomial architecture that approximates differential equations is proposed. The Legendre polynomial is chosen as a basis
in \cite{ref11}. But it should be noted, that in all these paper, the polynomial architectures are used as black box models, and the authors do not indicate its connection to the ODEs. Thus, the advantages of NN can only be partially exploited.

In all the described approaches, NN are trained to consider the initial conditions of the differential equations. This means that a NN has to be trained each time when the initial conditions are changed. The above-described techniques are applicable to the general form of differential equations but are able to provide only a particular solution of the system, that is a strict limitation of applicability.

Several types of Convolutional NN and Recurment NN have been developed in the last three years, cf. \cite{Ch19}. These deep networks corresponds to associated ODE discretization schemes. There is some benefit using these special network architecture, but still no clear explanation of it.

In this paper, we consider nonlinear systems of ODEs with polynomial right-hand side,
\begin{equation}
\label{odesystem}
\frac{d}{dt}\XX = \FF(t, \XX) = \sum_{k=0}^{\infty} P^{1k}(t)\XX^{[k]},
\end{equation}
where $t$ is an independent variable, $\XX \in R^n$ is a state vector, and $\XX^{[k]}$ means $k$-th Kronecker power of vector $\XX$. For example, for $\XX = (x_1, x_2)$ we have $\XX^{[2]} = (x_1^2, x_1x_2, x_2^2)$, $\XX^{[3]} = (x_1^3, x_1^2x_2, x_1x_2^2$, $x_2^3)$ after reduction of the same terms.

Such nonlinear systems arise in different fields such as automated control, robotics, mechanical and biological systems, chemical reactions, drug development, molecular dynamics, and so on. Moreover, often it is possible to transform a nonlinear equation (either ODE or PDE) to a polynomial form.

Based on the Taylor mapping technique, it is possible to build a polynomial neural network (PNN) that approximates the general solution of  \eqref{odesystem}. The weights of the PNN should be calculated at once directly from the system of ODEs. By a Taylor map we mean transformation $\mathcal M : \XX_0=\XX(t_0)\rightarrow \XX(t_1)$ in form of
\begin{equation}
	\label{tmap}
	\XX(t_1) = W_0 + W_1\,\XX_0+W_2\,\XX_0^{[2]}+\ldots+W_k\,\XX_0^{[k]},
\end{equation}
where $\XX, \YY \in R^n$, and matrices $W_i$ are weights. The ransformation \eqref{tmap} is linear in weights $W_i$ and nonlinear with respect to the $\XX_0$.

In literature, this map $\mathcal M$ can be referred to as Taylor maps and models \cite{ref_tm1}, tensor decomposition \cite{ref_tm2}, matrix Lie transform \cite{ref16}, exponential machines \cite{ref201}, and others. In fact, the transformation \eqref{tmap} is just a polynomial regression with respect to the components of $\XX$.

Though many numerical solvers for \eqref{odesystem} can be considered as maps, they commonly based on small time steps $\Delta t$ and weight matrices $W_i = W_i(\Delta t, \XX_0)$ that depend on $\XX_0$. In this paper, by mapping approach we mean transformation \eqref{tmap} with the weight matrices  $W_i = W_i(t_1-t_0)$ that are estimated for a large enough time interval $t_1-t_0 > \Delta t$ and do not depend on $\XX_0$. The greatest advantage of the mapping approach is computational performance. Instead of step-by-step integrating with numerical solvers, one can apply a map \eqref{tmap} that estimates dynamics in a large time interval for different initial conditions $\XX_0$ at the same accuracy.

\begin{figure}
\label{fig:1}
\centerline{
\includegraphics[width=0.4\textwidth]{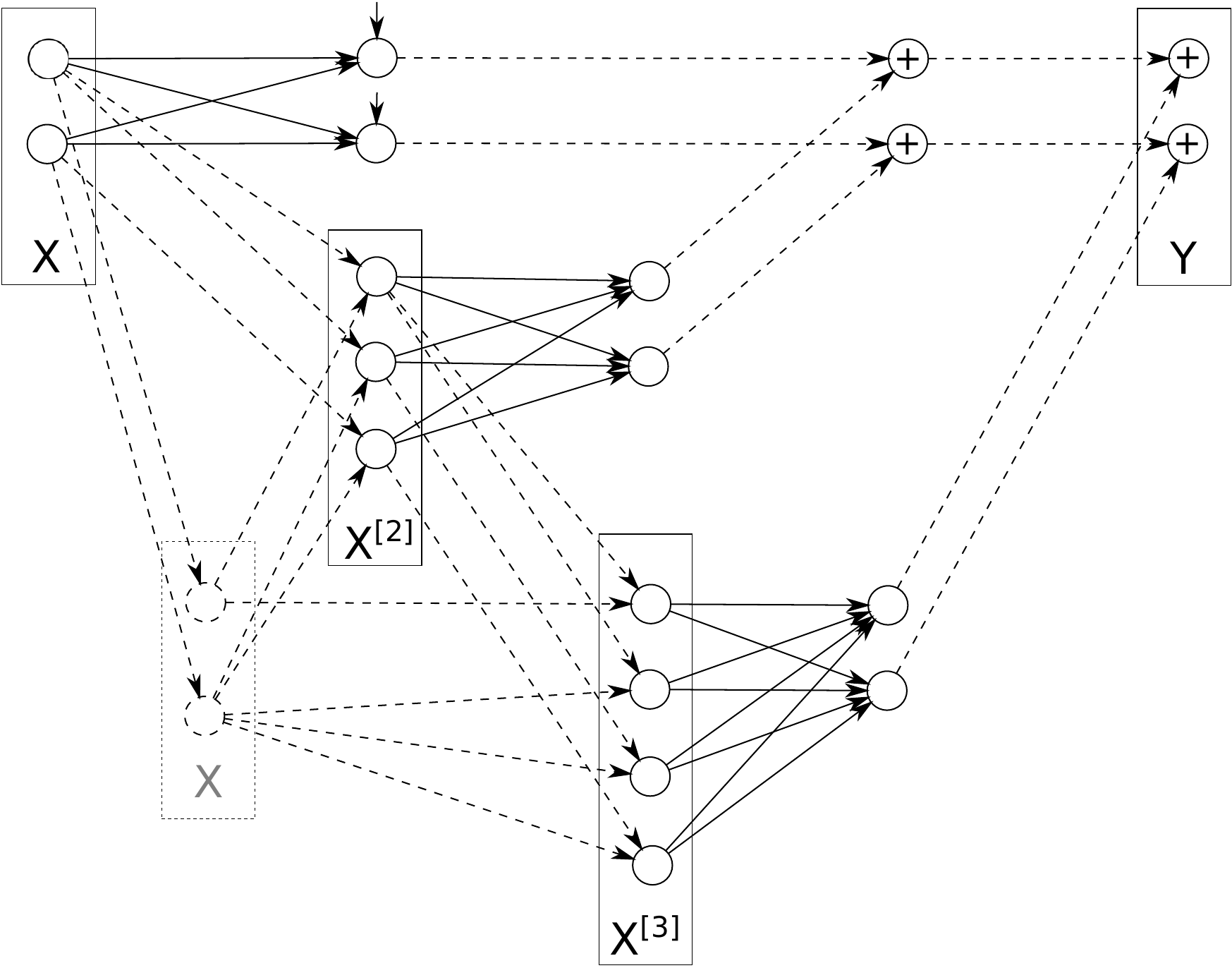}}
\caption{Polynomial neuron of 3rd order of nonlinearity}
\end{figure}

In the paper, we briefly consider an algorithm for calculating weight matrices $W_i$ in \eqref{tmap} for an arbitrary time step $t_1-t_0$. Considering map \eqref{tmap} as a neuron (Fig.~\ref{fig:1}), it is possible to design a PNN that simulates the dynamics of the given differential equation. While the PNN is connected to differential equations, it is also possible to use this architecture for data-driven identification of physical systems.

The rest of the paper is organised as follows. Sec.~2 describes the general approach for building Taylor maps for the systems of ODEs. Sec.~\ref{sec:simulation} corresponds to the PNNs weights calculating for dynamics simulation of applied physical systems. Sec.~\ref{sec:4000} is devoted to training PNN. The regularization method along with the examples of data-driven dynamics learning without equation consideration are described.

\section{Taylor maps for the system of ODEs}
\label{2001}
The solution of \eqref{odesystem} with the initial condition $\XX(t_0) = \XX_0$ during the time interval $t-t_0$ in its convergence region can be presented in the power series, cf. \cite{ref15,ref16},
\begin{equation}
\label{Lie}
\XX(t) =  \mathcal M(t-t_0) \circ \XX_0 =  \sum_{k=0}^{\infty} M^{1k}(t)\XX_0^{[k]},
\end{equation}
Theoretical estimations of accuracy and convergence of the truncated series in solving of ODEs can be found in \cite{ref17}. In \cite{ref14}, it is shown how to calculate matrices $M^{1k}$ by introducing new matrices $P^{ij}$.
The main idea is replacing
\eqref{odesystem} by the equation
\begin{equation}
\label{Lie_map}
\frac{d}{dt} M^{ik}(t) = \sum_{j=i}^{k} P^{ij}(t)M^{jk}(t),\;1\leq i < k.
\end{equation}
The last equation does not depend on $\XX_0$ and should be solved at once with initial condition $M^{kk}(t_0) = I^{[k]},\;M^{jk}(t_0) = 0, j\neq k$, where $I$ is the identity matrix. The truncated solution  of \eqref{Lie_map} for desirable time interval $t_1-t_0$ yields Taylor map \eqref{tmap} with $W_i = M^{1i}$.

One of the advantages of the described approach is simulation performance. Indeed, instead of step-by-step integrating of the equation \eqref{odesystem} with a small time step $\Delta t$ every time when the initial condition $\XX_0$ is changed, one should integrate map $\mathcal M(t)$ at once for the unique initial condition and the whole desirable time interval $t_1-t_0$. Then the same map can be applied for different initial conditions.

\section{Simulation of dynamical systems}
Since the Taylor mapping approach is commonly used in accelerator physics \cite{ref_12,ref25}, we demonstrate the proposed method with the simplified example of charged particle motion. We also introduce deep PNN for simulation and control of charged particle beam dynamics, and discuss a shallow PNN architecture for simulation of a stiff ODE.

\label{sec:simulation}
\subsection{Charged particle dynamics}
\label{3001}
The particle dynamics in the electromagnetic fields can be described by a system of ODEs that has a complex nonlinear form. For simplicity, let's consider an approximation \cite{ref202} of particle motion in cylindrical deflector written in form of \eqref{odesystem}
\begin{equation}
\label{cyl_defl}
\begin{split}
x' &= y,\\
y' &= -2x + x^2/R,
\end{split}
\end{equation}
where $R$ is equilibrium radius of particle bending, $x$ is deviation from this radius, and $x' = y$ is derivative on bending angle.

For an example, let's consider a deflector with $R =\SI{10}{m}$ that rotates a reference particle with initial conditions $x=0,\,x'=0$ on angle $\pi/4$. For simulation, we investigate dynamics of particle with nontrivial initial conditions that lead to particle oscillation.

Traditional approach to solve system \eqref{cyl_defl} is step-by-step integrating with numerical solvers. For this purpose, we use Runge--Kutta method of 4th order with fixed time step. To control the numerical error in this example, we use integrating angle in 30 times smaller than bending angle. Bigger steps introduce nonphysical dissipation in particle motion.
\begin{figure}
\centerline{
\includegraphics[width=0.35\textwidth]{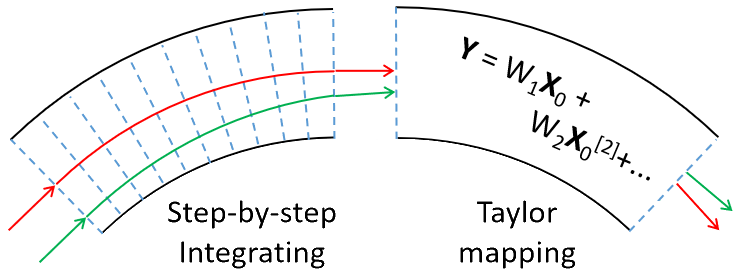}
}
\caption{Step-by-step numerical integrating and Taylor mapping}
\label{fig:2}
\end{figure}
In step-by-step integrating, to track particle inside the deflector, one has to do 30 steps. In contrast to this, mapping technique allows to calculate output state in single step (see Fig.~\ref{fig:2}).

Let's fine a 3rd order Taylor map \eqref{tmap} that transform initial particle state $\XX_0 = (x_0;y_0)$ at the entrance of the deflector to the resulting state $\XX_1 = (x_1;y_1)$ at the end of it,
\begin{equation}
\begin{split}
   \begin{pmatrix}
        x\\
        y
        \end{pmatrix}&=
        W_1
        \begin{pmatrix}
        x_0\\
        y_0
        \end{pmatrix}+
        W_2
        \begin{pmatrix}
        x_0^2\\
        x_0y_0\\
        y_0^2
        \end{pmatrix}+
        W_3
        \begin{pmatrix}
        x_0^3\\
        x_0^2y_0\\
        x_0y_0^2\\
        y_0^3
        \end{pmatrix}
\end{split}
\end{equation}
Combining this map with \eqref{cyl_defl}, one can write
\begin{equation}
    \begin{split}
        \begin{pmatrix}
        x\\
        y
        \end{pmatrix}'&=
        W_1'
        \begin{pmatrix}
        x_0\\
        y_0
        \end{pmatrix}+
        W_2'
        \begin{pmatrix}
        x_0^2\\
        x_0y_0\\
        y_0^2
        \end{pmatrix}+
        W_3'
        \begin{pmatrix}
        x_0^3\\
        x_0^2y_0\\
        x_0y_0^2\\
        y_0^3
        \end{pmatrix},\\
    \begin{pmatrix}
        x\\
        y
    \end{pmatrix}'&=
    \begin{pmatrix}
        0&1\\
        -2&0
    \end{pmatrix}
    \begin{pmatrix}
        x\\
        y
    \end{pmatrix}+
    \begin{pmatrix}
        0&0&0\\
        1/R&0&0
    \end{pmatrix}
    \begin{pmatrix}
        x^2\\
        xy\\
        y^2
    \end{pmatrix}.
    \end{split}
\end{equation}

Grouping the like terms of powers of $x_0$ and $y_0$ in the last system, one can obtain a system of ODEs that do not depends on $\XX_0 = (x_0; y_0)$ and represents dynamics of weight matrices
\begin{equation}
\begin{split}
W_1' &= f_1(W_1, W_2, W_3),\;\;\; W_1(0) = I,\\
W_2' &= f_2(W_1, W_2, W_3),\;\;\; W_2(0) = 0,\\
W_3' &= f_3(W_1, W_2, W_3),\;\;\; W_3(0) = 0,\\
\end{split}
\end{equation}
where $f_i$ is functions arising after like terms grouping. By integrating this system during the interval $[0; \pi/4]$, we can receive the desired map. For example, the map up to two digits is
\begin{equation*}
\begin{split}
        \begin{pmatrix}
        x_1\\
        y_1
    \end{pmatrix}&=
    \begin{pmatrix}
        0.44& 0.63\\
        -0.13\cdot10&0.44
    \end{pmatrix}
    \begin{pmatrix}
        x_0\\
        y_0
    \end{pmatrix}+\\
    &\begin{pmatrix}
        0.23\cdot10^{-1}& 0.12\cdot10^{-1}& 0.26\cdot10^{-2}\\
        0.40\cdot10^{-1}& 0.35\cdot10^{-1}& 0.12\cdot10^{-1}
    \end{pmatrix}
    \begin{pmatrix}
        x_0^2\\
        x_0y_0\\
        y_0^2
    \end{pmatrix}+\\
    &\begin{pmatrix}
        0.21\cdot10^{-3} 0.17\cdot10^{-3} 0.47\cdot10^{-4} 0.56\cdot10^{-5}\\
        0.83\cdot10^{-3} 0.95\cdot10^{-3} 0.32\cdot10^{-3} 0.47\cdot10^{-4}
    \end{pmatrix}
    \begin{pmatrix}
        x_0^3\\
        x_0^2y_0\\
        x_0y_0^2\\
        y_0^3
    \end{pmatrix}.
\end{split}
\end{equation*}

Using this polynomial transformation, one can calculate state of the particle at the end of the deflector for arbitrary initial conditions $\XX_0 = (x_0; y_0)$. The simulation in this case is 160 times faster than Runge--Kutta based simulation.

\begin{figure}
\centerline{
\includegraphics[width=0.45\textwidth]{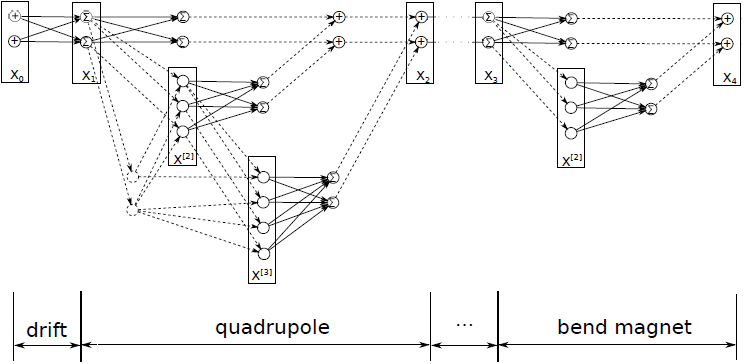}
}
\caption{Deep PNN for simulation of a charged particle accelerator}
\label{fig:3}
\end{figure}

For long-term charged particle dynamics investigation, the traditional step-by-step numerical methods are not suitable because of the performance limitation. Instead of solving differential equations directly, one can estimate Taylor maps of high orders of nonlinearity for each control element in an accelerator. By combining such maps consequently, one can obtain a deep polynomial neural network that represents the whole accelerator lattice (see Fig.~\ref{fig:3}). In such PNN each layer is represented by a Taylor map and corresponds to the physical element in charge particle accelerator.

The described approach was used for simulation of nonlinear spin-orbit dynamics in EDM (electric dipole moment) search project \cite{ref25,ref_12}. The dynamics of particles was described by nine-dimensional state vector. The deep PNN architecture had more then 100 polynomial layers that represent lattice of the accelerator. The computational performance was increased in 1500 times in comparison with traditional step-by-step integrating.

Along with the computational performance, another advantage of the proposed model is uncertainties accounting. It is easy to represent misalignments and field errors in physical accelerator by introducing additional polynomial layers inside the PNN. This feature may be essential for control of accelerators, while the PNN architecture provides a possibility to fit layers with measured data.

\subsection{The Rayleight-Plesset equation}
\label{sec:3002}
The Rayleight-Plesset equation governs the dynamics of a spherical gas bubble in an infinite body of in-compressible liquid. It is derived using the conservation of mass and momentum which leads to a highly non-linear second order ODE for the bubble radius $R$.
 The pressure inside the bubble is denoted by $p_B$ and the pressure outside far a way from the bubble is denoted by $p_{\infty}$,  $\rho$ is the density of the surrounding liquid, assumed to be constant. The equation of motion has the form
\begin{equation}\label{eq:2000}
R \frac{ d^2 R}{d t^2}  + \frac{3}{2} R \left(\frac{d R}{d t}\right)^2 = \frac{1}{\rho} (p_B-p) \,.
\end{equation}
Viscous terms as well as surface tension are neglected. Provided that $p_B(t)$ is known and $p_{\infty}(t)$ is given, the Rayleigh–Plesset equation can be used to solve for the time-varying gas bubble radius $R(t)$ which includes collapses $(R\to 0)$ and rebounds.

Because we study the numerical solvers for the Rayleigh-Plesset equation, we assume that the pressure difference $p_B-p_{\infty}$ is constant, i.e. $p_B=\SI{2300}{Pa}$ and $p_{\infty}=\SI{1e5}{Pa}$ for simplicity. The solution contains very strong gradients during the collapse phase and the following rebound phase, i.e it is a stiff ODE. In order to guarantee the accuracy of the numerical solution, very small time steps as well as a high order of the solver must be used.    The Rayleigh-Plesset equation \eqref{eq:2000} is reformulated in a system of first order ODE's with polynomial right-hand side
\begin{equation}
\begin{split}
 \frac{d y_1}{d t} &=  y_2,\\
\frac{d y_2}{d t} &= -\frac{(p_B-p)}{\rho}y_3 -\frac{3}{2} y_2^2,\\
\frac{d y_3}{d t} &= -y_3^2y_2,
\end{split}
\end{equation}
where $y_1=R$, $\dot{R}=y_2$, $y_3=1/R$, and
the initial conditions are $y_1(0)=R(t=0)=R_0$, $y_2(0)=0$, and $y_3(0) = 1/y_1(0)$.

\begin{figure}[h]
\centering
\includegraphics[width=0.35\textwidth]{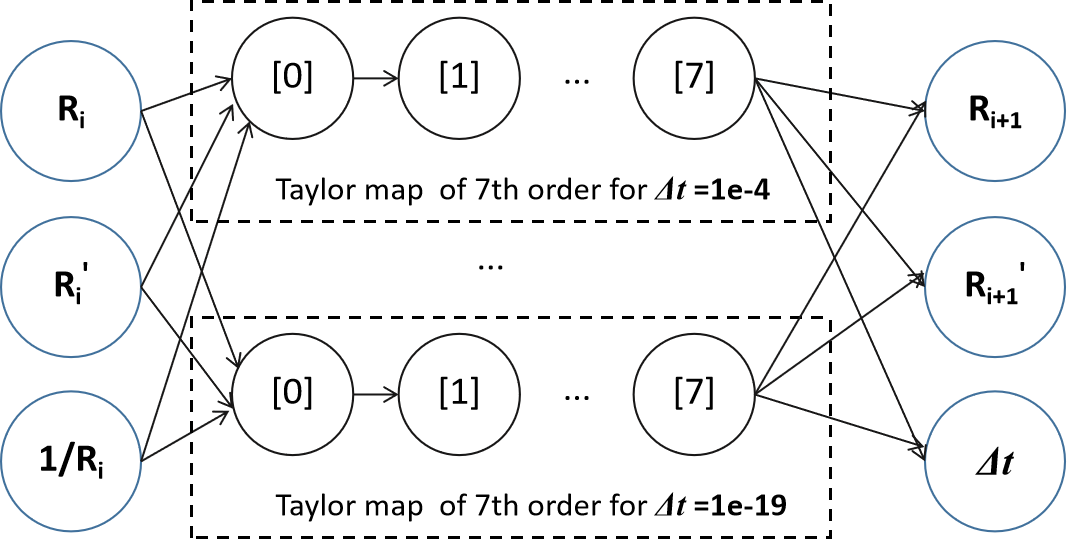}
\caption{Adaptive mapping technique in a shallow PNN architecture}
\label{fig:4}
\end{figure}

\begin{figure}[h]
\centering
\includegraphics[width=0.40\textwidth]{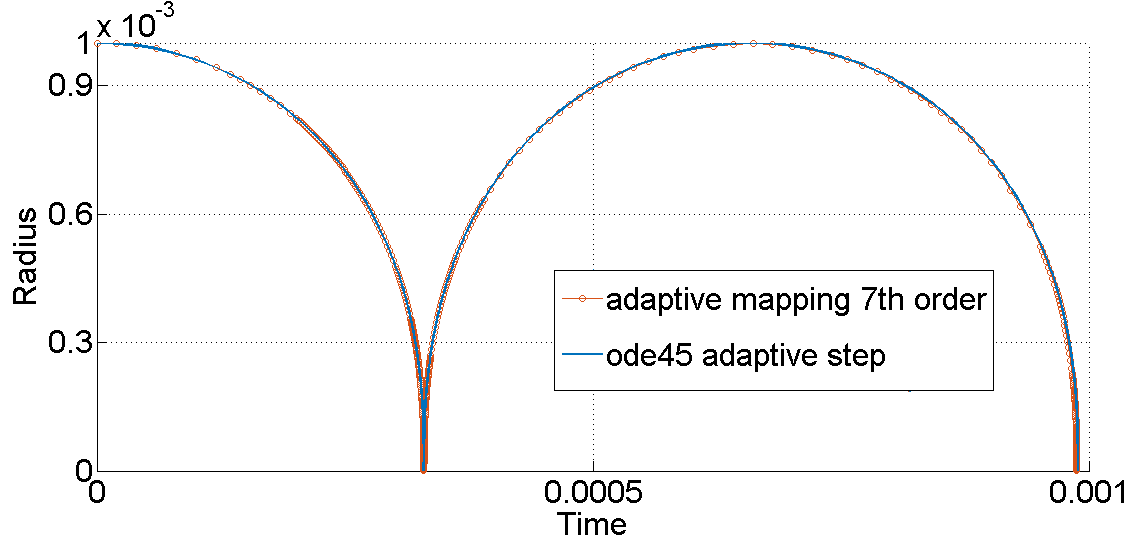}
\caption{Simulation of the Rayleight-Plesset equation: blue line for ode45, orange circles for PNN}
\label{fig:5}
\end{figure}

Since the equation is a stiff ODE, we use an adaptive mapping approach. This means that we built Taylor maps of 7th order for different time intervals ranging from $\Delta t = \SI{1e-4}{s}$ to $\Delta t = \SI{1e-19}{s}$. Then we apply maps based on the relative error during the simulation. For this approach, the polynomial neurons can be organized in a shallow architecture that provides possibility to simulate of a bubble motion with control of accuracy. To compare the computational performance we run simulation for three initial conditions $R_0 \in \{\SI{0.85e-3}{m}, \SI{1e-3}{m}, \SI{1.15e-3}{m} \}, \dot R_0 = 0.$ up to the collapsing time. The adaptive mapping approach is 2.5 times faster then the solver for stiff differential equations ode45 from \cite{Matlab:2018} with similar accuracy (see Fig.~\ref{fig:4}, \ref{fig:5}).

\subsection{The Burgers' equation}
\label{sec:3001}
Burgers' equation is a fundamental partial differential equation that occurs in various areas, such as fluid mechanics, nonlinear acoustics, gas dynamics, and traffic flow. This equation is also often used as a benchmark for numerical methods. In \cite{ref20}, a feed-forward neural network is trained to satisfy Burgers' equation and certain initial conditions, but the computational performance of the approach is not estimated. In this section, we demonstrate how to build a PNN that solves Burgers' equation and does not require training to satisfy different initial conditions.

Burgers' equation has a form
\begin{equation}
\label{burgers_de}
    \frac{\partial u(t,x)}{\partial t} + u(t,x)\frac{\partial u(t,x)}{\partial x} = \nu \frac{\partial^2 u(t,x)}{\partial x^2}.
\end{equation}

\noindent
Following \cite{ref19} for benchmarking, we use an analytic solution
\begin{equation*}
\begin{split}
    u_1(t,x) &= -2\frac{\nu}{\phi(t,x)}\frac{d\phi}{dx}+4,\\
    \phi(t,x) &= exp\frac{-(x-4t)^2}{4\nu(t+1)} + exp\frac{-(x-4t-2\pi)^2}{4\nu(t+1)},
\end{split}
\end{equation*}
and a traditional numerical method
\begin{equation}
\label{FDM}
    \frac{u^{n+1}_i - u^{n}_i}{\Delta t} + u^{n}_i\frac{u^{n}_i - u^{n}_{i-1}}{\Delta  x} = \nu \frac{u^n_{i+1} - 2u^n_i + u^n_{i-1}}{\Delta x^2},
\end{equation}
where $n$ stands for the time step, and $i$ stands for the grid node.

The equation (\ref{FDM}) presents a finite difference method (FDM) that consists of an Euler explicit time discretization scheme for the temporal derivatives, an upwind first-order scheme for the nonlinear term, and finally a centered second-order scheme for the diffusion term. The time step for benchmarking is fixed to $\Delta t = \SI{2.5e-4}{s}$ with the uniform spacing of $\Delta x = 2\pi/1000$, $\nu =\SI{0.05}{m^2/s}$. Thus, for the numerical solution for times from $t = \SI{0}{s}$ to $t = \SI{0.5}{s}$ on $x\in[0, 2\pi]$, the method requires the mesh with 1000 steps on space coordinate $x$ and 2000 time steps.

\begin{figure}[h!]
\centering
\includegraphics[width=0.4\textwidth]{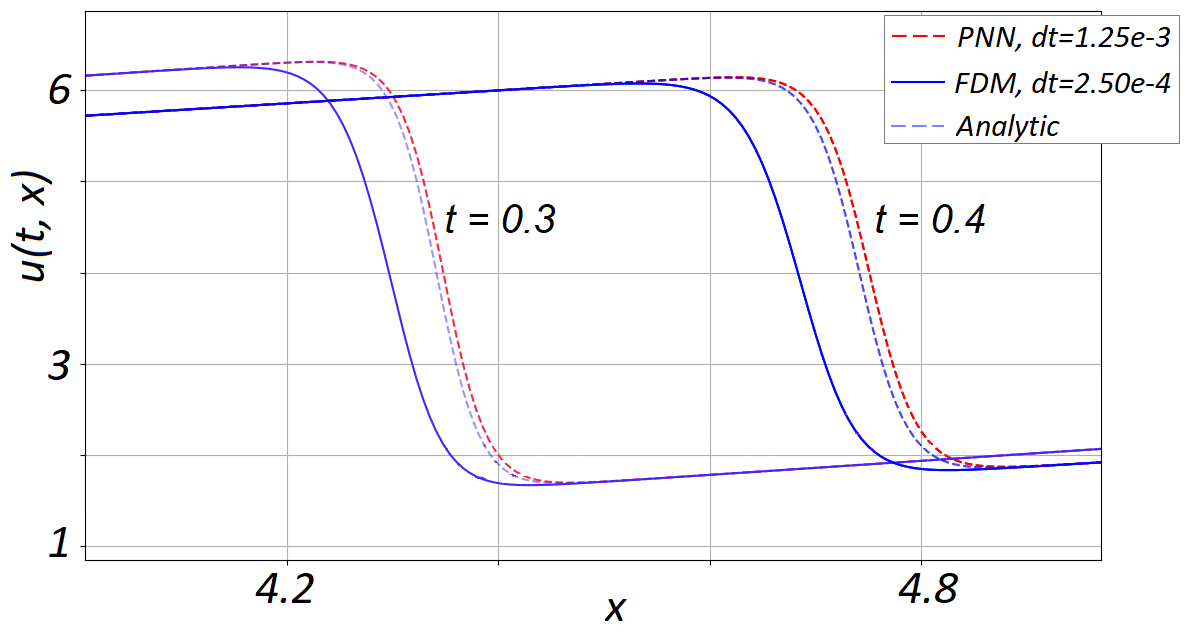}
\caption{Burgers' equation: blue line for FDM, red dashed line for PNN, blue dashed line for analytic solution}
\label{fig:fig6}
\end{figure}

It is indicated in \cite{ref19} that the FDM introduces a dispersion error in the solution (see Fig.~\ref{fig:fig6}). Such error can be reduced by increasing the mesh resolution, but then the time step should be decreased to respect the stability constraints of the numerical scheme.

To build a PNN for solving the Burger's equation we translate the eq. \eqref{burgers_de} to a system of ODEs and build a Taylor map in accordance with Sec.~\ref{2001}. Assuming that the right-hand side of the eq. \eqref{burgers_de} can be approximated by a function $f(x, u(t,x))$ and considering this approximated equation as a hyperbolic one, it is possible to derive the system of ODEs

\begin{equation}
    \label{burgers_ode}
    \frac{d}{dt}
    \begin{pmatrix}
    \XX\\
    \UU
    \end{pmatrix}
    =
    \begin{pmatrix}
    \UU\\
    f(x, \UU)
    \end{pmatrix},
\end{equation}
where $\UU=(u_1, \ldots, u_{1000})$, $u_i(t) = u(t, x_i)$, and $\XX=(x_1, \ldots, x_{1000})$ is vector of discrete stamps on space. This transformation from PDE to ODE is well known and can be derived using the method of characteristics and direct method \cite{ref23}. If $f(x, u(t,x))$ is the same discretization as in (\ref{FDM}), then the equation (\ref{burgers_ode}) leads to the system of 2000 ODEs
\begin{equation*}
    \begin{aligned}
    x'_i &= u_i,\;\;\;\;
    u'_i &= f(\nu, u_{i+1}, u_i, u_{i-1}, x_{i+1}, x_i, x_{i-1}),
    \end{aligned}
\end{equation*}
which can be easily expanded to the power series with respect to the $\XX$ and $\UU$ up to the necessary order of nonlinearity.

Since there are only first and second order approximations in the benchmarking FDM scheme, we built a Taylor map of only the first order in a time interval $\Delta t =\SI{1.25e-3}{s}$. This time step is five times larger than that used in the benchmarks. The numerical solution provided by the resulting PNN is presented in Fig.~\ref{fig:fig6}, and the accuracy and performance are compared in Tab.~\ref{table2}.

\begin{table}
\centering
\caption{Comparison of the simulation of the Burgers equation by FDM and PNN}\label{table2}
\begin{tabular}{|l|c|c|c|}
\hline
\textbf{Method} & \textbf{Time} & \textbf{Elapsed} & \textbf{MSE for} \\

 & \textbf{step} & \textbf{time} & $\mathbf{u_1(0.5,x)}$ \\
\hline
FDM&\;\;\;\;$\SI{2.50e-4}{s}$\;\;\;\;&\;\;\;\;$\SI{0.055}{s}$\;\;\;\;&\;\;\;\;$8.0\cdot10^{-2}$\;\;\;\;\\
\hline
PNN&$\SI{1.25e-3}{s}$&$\SI{0.016}{s}$ & $5.5\cdot10^{-3}$ \\
\hline
\end{tabular}
\end{table}

\noindent

The PNN numerically estimates dynamics in a larger time interval and provides better accuracy with less computational time in comparison with FDM of the same order of derivative approximations. If the FDM scheme is adjusted to a higher accuracy, the computational time will be increased even more. Accuracy is calculated as the MSE metric between the numerical solution and its analytic form at final time $t =\SI{0.5}{s}$.

Since the derived from the differential equation PNN is an approximation of the general solution, it can be used for simulation of the dynamics with different initial conditions without weight modification. For example, for another analytic solution
$$
u_2(t, x) = (1+exp(0.5(x-0.5 t)/\nu)^{-1},
$$
the same PNN provides a numerical solution in time $t=\SI{0.5}{s}$ with MSE error $\SI{7.2e-8}{s}$, while FDM yiels $\SI{1.2e-7}{s}$. The elapsed times are the same
since the mesh sizes were not changed.

\begin{figure}
\centering
\includegraphics[width=0.36\textwidth]{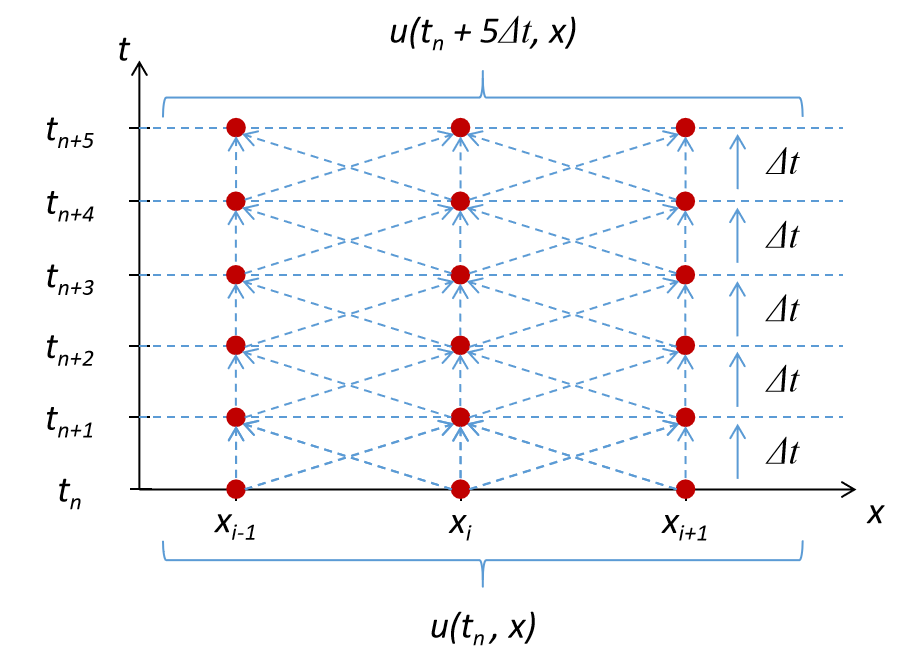}\\
\includegraphics[width=0.36\textwidth]{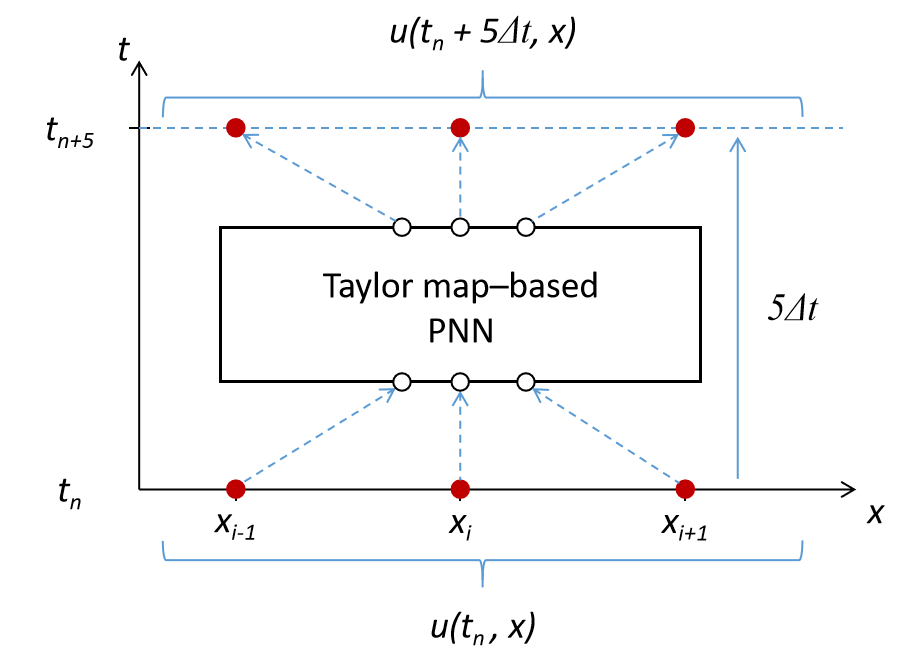}
\caption{Numerical schemes for FDM, $\Delta t = 2.5-4 s$ (top Fig.) and PNN, $\Delta t =\SI{1.25e-3}{s}$ for (\ref{burgers_de})}
\label{fig6}
\end{figure}

\section{Learning dynamical systems from data}
\label{sec:4000}
In this section, we discuss training of the PNN. We demonstrate how one can adjust weights of the PNN for certain initial conditions of the Burgers' equation by additional training of the PNN derived from the equation. We also introduce a regularization of the PNN based on combinatorial binary optimization and demonstrate the proposed methods in a practical application of a data-driven system learning.
\subsection{Training PNN for certain initial conditions}
\label{sec:4001}
The PNN derived from differential equation works for any initial condition with the same level of accuracy. The accuracy depends on the size of the PNN, namely the order of non-linearities in the map (\ref{tmap}). But it is still possible to train PNN for certain initial conditions. This fitting increases accuracy only for this particular solution, but probably decrease accuracy for others.

In order to train PNN for the Burgers equations, we consider deep architecture with 400 layers with shared weights. Each layer M is defined by Taylor map (\ref{tmap}) for time $dt=\SI{1.25e-3}{s}$:
$$
    u(t=0, x) \rightarrow M \rightarrow \ldots \rightarrow M \rightarrow u(t=0.5, x).
$$
The loss function is defined as the inconsistency of the numerical solution to differential equations. The initial values of weights are calculated from the PDE with corresponding Taylor map.

Initially, the PNN has $4\cdot10^{6}$ weights with only $5000$ of them that are not equal to zero. The point of training is to adjust these weights to achieve better accuracy for the given initial conditions. We implemented a simplified training based on coordinate descent method that allows to increase accuracy in 3 times, but in the same time requires lots of computational time.

\subsection{Regularization of the PNN}
\label{sec:4002}
It is common for physical systems, that map is presented by sparse matrices $W_i$. So it is important to identify non-zero weight in the polynomial neural network. Classical $L_1$ and $L_2$ regularization terms do not work in this case since they tend to decrease weight. Weights in PNN can be large by absolute value but the total amount of non-zero weights should be as small as possible.
To solve this issue, one can introduce a binary mask $B_i$ that is applied for weight matrices,
\begin{equation}
\label{QUBO_reg}
\XX_{i+1} = (B_0*W_0) + (B_1*W_1)\XX_i+ (B_2*W_2)\XX_i^{[2]}+\ldots,
\end{equation}
where $(B_i*W_i)$ means element-wise multiplication.

During the fitting of weight matrices $W_i$, it is necessary also to find an optimal mask $B_i$. If the weights $W_i$ are fixed after each training epoch, the map \eqref{QUBO_reg} becomes linear for the components of $B_i$.

\begin{figure}
\centering
\includegraphics[width=0.45\textwidth]{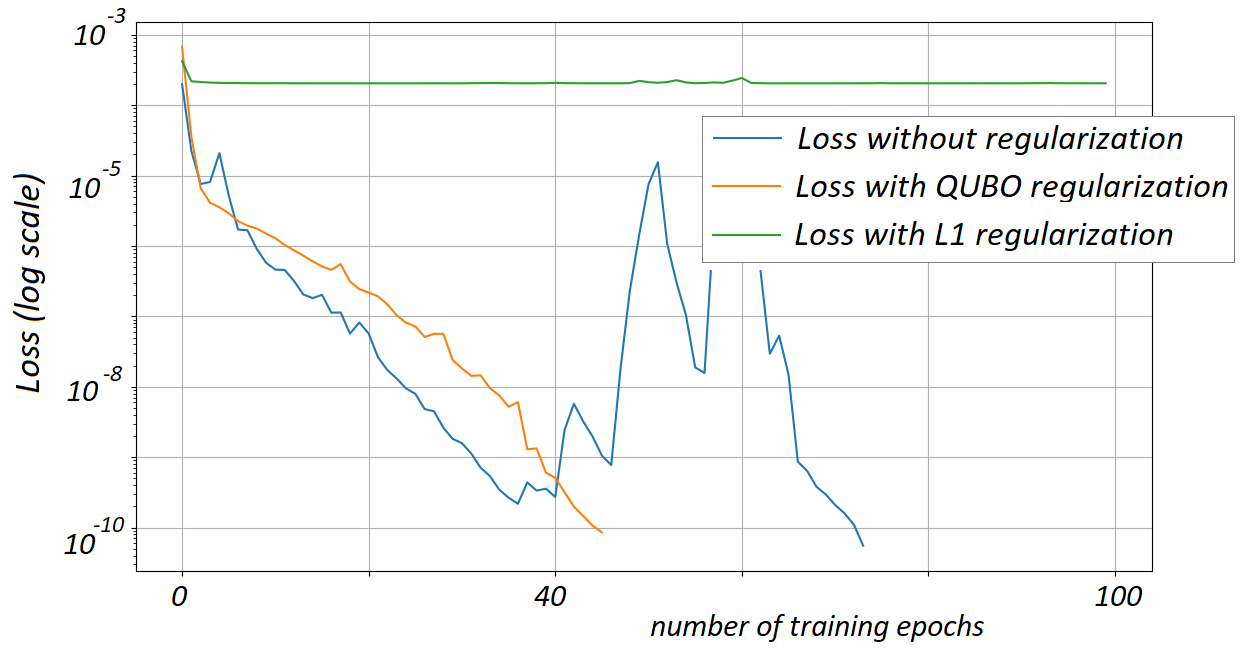}
\caption{Loss functions and number of epochs for training PNN for the Van der Pol oscillator: without regularization (blue line), with L1 regularization (green line), and with proposed QUBO--based regularization.} \label{fig501}
\end{figure}

It is possible to formulate regularization as a high-order binary optimization problem and translate it into a quadratic binary minimization problem (QUBO) that can be run on Quantum Annealers. For example,the loss function $||X_{i+1} - (\mathcal B*\mathcal M)\circ\XX_i||$ translates directly to QUBO, while loss function that introduced for Burgers equation translates to high-order unconstrained binary optimization problem. By introducing additional binary variables it is possible finally to formulate regularization as QUBO.

To demonstrate the advantage of QUBO--based regularization, we implemented a simplified example with the Van der Pole oscillator. The equation is widely used in the physical sciences and engineering and can be used for the description of the pneumatic hammer, steam engine, periodic occurrence of epidemics, economic crises, depressions, and heartbeat. The equation has well-studied dynamics and is widely used for testing of numerical methods, e.g. \cite{ref18}.

The Van der Pol oscillator is defined as the system of ODEs $x'' = x' - x - x^2x'$ that can be presented in the form of
\begin{equation}
\label{vdp_ode}
    \begin{aligned}
        x' &= y,\\
        y' &= y-x-x^2y.
    \end{aligned}
\end{equation}

We generate a training data as a particular solution $\{\XX_i\}_{i=1;n}$ of the system \eqref{vdp_ode} with the initial condition $\XX_0 = (-2, 4)$. After this solution was generated, the equation is not used further.

\begin{figure}
\centering
\includegraphics[width=0.38\textwidth]{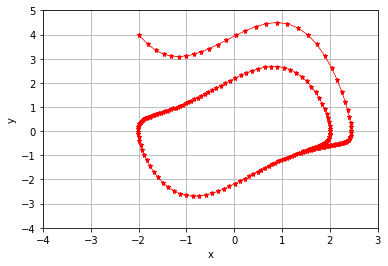}
\caption{Training data for the neural network as a numerically generated particular solution.}
\label{fig301}
\end{figure}

Having this training data set (Fig.~\ref{fig301} ), the proposed neural network can be fitted with the mean squared error (MSE) as a loss function based on the norm
$$
||\XX_{i+1} - \mathcal M\circ \XX_i ||=
||
\begin{pmatrix}
x_{i+1}\\
y_{i+1}\\
\end{pmatrix}
-W_0 - \ldots
-W_3
\begin{pmatrix}
x_i^3\\
x_i^2y_i\\
x_iy_i^2\\
y_i^3
\end{pmatrix}
||.
$$

We implemented the above-described technique in Keras/TensorFlow and fitted a third-order Lie transform--based neural network with an Adamax optimizer. After each training epoch we applied QUBO--based regularization for the weights of the PNN.

Fig.~\ref{fig501} shows that QUBO--based regularization can significantly decrease the number of epochs required to achieve an appropriate level of accuracy. However, the time required to solve QUBO problem is longer then training without regularization. It should be noted that using of special hardware (e.g. quantum or digital annealers) instead of classical QUBO-solvers can potentially provide additional improvement in calculation time.

\begin{figure}
\centering
\includegraphics[width=0.38\textwidth]{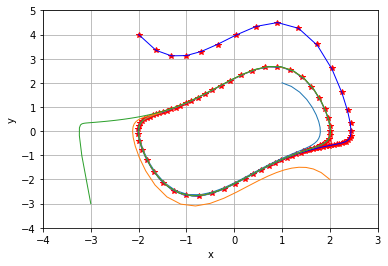}
\caption{Prediction for new initial condition that were not presented during training.}
\label{fig302}
\end{figure}

The generalization property of the network can be investigated by examining prediction not only from the training data set but also for new initial conditions. Fig.~\ref{fig302} demonstrates predictions that are calculated starting also at the new points $(1, 2), (2, -2)$, and $(-3, -3)$ that were not presented to the PNN during fitting. For the prediction starting from the training initial condition, the mean relative error of the predictions is $4.8\cdot10^{-5}$. For the new initial conditions, the mean error is even less and equals to $9.8\cdot 10^{-6}$.

\subsection{Data-driven identification of a zinc-air energy storage system}
\label{sec:4003}
Zinc-air batteries are one of the most promising energy storage systems. An effective mathematical model is a key part of a battery management system, responsible for its operation control and internal state evaluation. The model-based analysis is an effective tool in the design and manufacture of power supplies, as well as in the analysis of their behavior under different operating conditions, cf. \cite{Lao-atiman2019}.

There are several approaches for building dynamic model of a battery.
By now the researches were mostly focused on the zinc-air batteries mathematical models derivation based on electrochemical principles of its functioning. This implies utilization of complicated nonlinear PDEs, describing the battery behavior precisely, but requiring time-consuming numerical procedures for their solution.

Another type is a “black box” model or data-driven model that is formed by learning on experimental data using artificial neural networks. Since this approach does not imply the presence of state equations in the model, it  works only on similar data to what the NN was trained on. This requires  availability  of operational data from real environment of sufficient volume and quality, which in practice cannot always be guaranteed.

Thus, state space models with lumped parameters are more suitable for analysis, control, and optimization of power supplies, since in this case, it includes battery general characteristics, like voltage, current, state of charge and so on. In the most general form without specification of any details, it can be presented as
\begin{equation}\label{eq.battery}
\begin{array}{rcl}
x(k+1)&=&f(x(k),u(k)),\\
y(k)&=&g(x(k),u(k)).
\end{array}
\end{equation}

It should be noted that real power sources exhibits strongly nonlinear effects, that are to be included as a nonlinear part of $f(\cdot)$ and $g(\cdot)$ functions of mathematical model \eqref{eq.battery} from \cite{olaru2019}.
This implies manual derivation of the equations for dynamics description and identification of its parameters. In  this  work  we  take  an  alternative  approach by data-driven construction of mathematical model from  a  set  of controlled   experiments \cite{Lao-Atiman2019168}.   We  investigate   whether   it   is possible   to   extract relevant   features   from   current   and voltage    measurements    collected    during battery    discharge  in  various fixed  loading  conditions.

\begin{figure}
\centering
\includegraphics[width=0.5\textwidth]{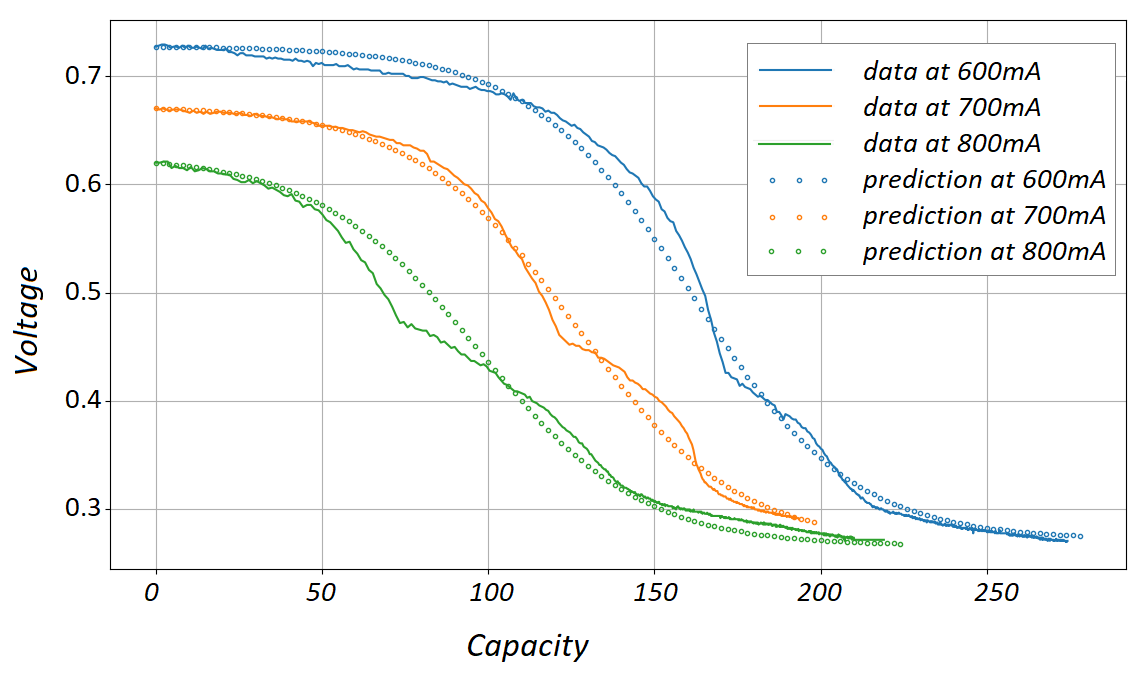}
\caption{The input data (lines) and predictions (dots) provided by PNN.}
\label{fig502}
\end{figure}

For demonstration of this approach, we consider only voltage dynamics for discharge currents  $\SI{600}{mA}$, $\SI{700}{mA}$ , and $\SI{800}{mA}$ and solve an identification problem only for regions where voltage decreases in time.

\begin{figure}
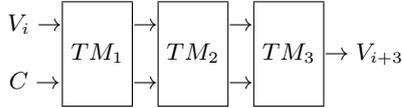

\centering
\begin{equation*}
    \begin{matrix}
    V_i\rightarrow\\
    \\
    C\rightarrow
    \end{matrix}
    \boxed{
    \begin{matrix}
    \\
    TM_1\\
    \\
    \end{matrix}
    }
    \begin{matrix}
    \rightarrow\\
    \\
    \rightarrow
    \end{matrix}
    \boxed{
    \begin{matrix}
    \\
    TM_2\\
    \\
    \end{matrix}
    }
    \begin{matrix}
    \rightarrow\\
    \\
    \rightarrow
    \end{matrix}
    \boxed{
    \begin{matrix}
    \\
    TM_3\\
    \\
    \end{matrix}
    }
    \begin{matrix}
    \\
    \rightarrow V_{i+3}\\
    \\
    \end{matrix}
\end{equation*}
\caption{PNN architecture for learning of voltage dynamics.}
\label{fig5020}
\end{figure}

Initial data taken from the open source \cite{Lao-Atiman2019168} were filtered with simple moving average window and aligned to constant time stamp of 1ms. To present dynamics of the voltage decreasing, we use polynomial neural network with three layers (see Fig.~\ref{fig5020}). Each layer is represented by a Taylor map, namely $TM_1$ and $TM_3$ are Taylor maps of second order, $TM_2$ is a Taylor map of fifth order.

To train the PNN, vector of voltage at the given time stamp and the value of discharge current is used as input data, and voltage after three time stamps is used as output data. Also, the described above regularization were used.

After training, the dynamics of the voltage decreasing is calculated as consequent prediction performed by PNN. Starting with initial voltage $V_0$ at time $t=0$ we calculate dynamics by formula
\begin{equation*}
V_0 \rightarrow V_1=PNN(V_0)\rightarrow \ldots\rightarrow V_k=PNN(V_{k-1}).
\end{equation*}

In this way, PNN plays role of the model of the dynamical system that can calculate dynamics with different initial conditions $V_0$.

Fig.~\ref{fig502} shows the results of the voltage dynamics simulation with the trained PNN. While additional features like derivatives of voltage or power can potentially increase the accuracy, the built PNN preserves physical properties of the dynamics. For example, the PNN successfully predicts the lower level of voltage that is the same for all currents in contrast to simple function approximation, that can give aberrant negative values of voltage, cf. \cite{olaru2019}.

One of the advantage of the polynomial architectures in comparison with other machine learning methods is its coincidence with theory of dynamical systems and differential equations. A PNN trained with data can provide mathematical model with physical properties preservation. For example, in \cite{ref1001} it is shown, that MLP and LSTM neural networks just memorized data even for simplified physical systems. They can not predict dynamics for data that have not been presented during fitting. PNN architecture allows to extrapolate dynamics on new initial condition and preserve physical properties of the dynamical systems.

\section{Supplementary code}
The program code that contains the considered examples can be found in the following GitHub repositories.

\vspace{0.5cm}

\noindent
\textbf{github.com/andiva/DeepLieNet}: implementation of PNN in Keras/TensorFlow  (sections \ref{sec:1001}, \ref{sec:3001}, \ref{sec:3002}, and \ref{sec:4003}):

\noindent
\textbf{/demo/Deflector}: cylindrical deflector simulation\\
\textbf{/demo/accelerator.ipynb}: storage ring simulation\\
\textbf{/demo/Ray\_Ples}: simulation of the Rayleight-Plesset equation
\\
\textbf{/demo/Battery}: trained PNN for voltage dynamics

\noindent
\textbf{github.com/andiva/AQCC}: PNN for simulation of the Burgers equation (Sec.~\ref{sec:3001}), training PNN for the Burgers equation (Sec.~\ref{sec:4001}), QUBO--based regularization and training of PNN for the Van der Pol oscillator (Sec.~\ref{sec:4002}).

\noindent

\section{Conclusion}
Since the Taylor maps and polynomial neural network have strong connection to systems of differential equations, these methods can be used for physical system simulation and data-driven identification. If the differential equations for the system are known, it is possible to calculate weights of the PNN with necessary level of accuracy. In this way, the built PNN can be used for simulation of the dynamics instead of the traditional numerical solvers. It is shown, that Taylor mapping and PNN provides the same or even better accuracy with less computational time in comparison to step-by-step integrating of the equations.

If the equations of the system are not known, the PNN can be trained by scratch with the available data. One can define an architecture of the PNN and fit weights directly with data. Instead of fitting of a parametrized equations, this approach allows to fit weights of the PNN directly.

In the article, we consider simulation of dynamical system arising in practical application with the PNN. For training PNN, we introduces QUBO--based regularization and demonstrated data-driven dynamics learning with both  simplified Van der Pol oscillator and applied problem of the batteries development.

We do not consideed questions of accuracy, PNN architecture selection, and optimization methods for training in details. This work should be done in further research. With the provided examples, we rather demonstrated the applicability of Taylor mapping methods that are commonly used in dynamical systems investigation to the theory of neural network and data-driven system learning.

\ack We would like to thank Prof. Dr. Sergei Andrianov for his help with explanation of Taylor mapping techniques for solving of system of differential equations.

\bibliography{ecai}

\end{document}